\def\year{2020}\relax
\documentclass[letterpaper]{article} 
\usepackage{aaai20}  
\usepackage{times}  
\usepackage{helvet} 
\usepackage{courier}  
\usepackage[hyphens]{url}  
\usepackage{graphicx} 
\usepackage{graphicx}  
\nocopyright 

\usepackage{latexsym}
\usepackage{wrapfig}
\usepackage{color}
\usepackage{amsfonts}
\usepackage{amsmath}
\usepackage{multirow}
\usepackage{booktabs}
\usepackage{comment}
\usepackage{subcaption}

\title{A Study of Multilingual Neural Machine Translation}

\author{Xu Tan, Yichong Leng, Jiale Chen, Yi Ren, Tao Qin, Tie-Yan Liu \\
Microsoft Research Asia\\
\{xuta,taoqin\}@microsoft.com
}

\begin{document}
\maketitle

\begin{abstract}
Multilingual neural machine translation (NMT) has recently been investigated from different aspects (e.g., pivot translation, zero-shot translation, fine-tuning, or training from scratch) and in different settings (e.g., rich resource and low resource, one-to-many, and many-to-one translation). This paper concentrates on a deep understanding of multilingual NMT and conducts a comprehensive study on a multilingual dataset with more than 20 languages. Our results show that (1) low-resource language pairs benefit much from multilingual training, while rich-resource language pairs may get hurt under limited model capacity and training with similar languages benefits more than dissimilar languages; (2) fine-tuning performs better than training from scratch in the one-to-many setting  while training from scratch performs better in the many-to-one setting; (3) the bottom layers of the encoder and top layers of the decoder capture more language specific information, and just fine-tuning these parts can achieve good accuracy for low-resource language pairs; (4) direct translation is better than pivot translation when the source language is similar to the target language (e.g., in the same language branch), even when the size of direct training data is much smaller; (5) given a fixed training data budget, it is better to introduce more languages into multilingual training for zero-shot translation.

\end{abstract}

\section{Introduction}
Neural machine translation (NMT) has witnessed rapid progress in recent years~\cite{DBLP:journals/corr/BahdanauCB14,DBLP:conf/emnlp/LuongPM15,sutskever2014sequence,DBLP:journals/corr/WuSCLNMKCGMKSJL16,DBLP:conf/icml/GehringAGYD17,shen2018dense,DBLP:conf/nips/VaswaniSPUJGKP17,he2018layer,DBLP:journals/corr/abs-1803-05567} for single-pair translation with a large number of bilingual sentence pairs. Given that there are over 7000 languages in the world\footnote{https://www.ethnologue.com/browse} and most of them are low-resource,  multilingual NMT~\cite{DBLP:conf/acl/DongWHYW15,DBLP:journals/corr/LuongLSVK15,DBLP:conf/naacl/FiratCB16,DBLP:journals/corr/abs-1804-08198,DBLP:journals/tacl/JohnsonSLKWCTVW17,DBLP:journals/corr/HaNW16} is proposed to handle multiple languages, with the benefit of reducing the number of models, simplifying model training, easing online maintenance cost, and enhancing low-resource translation. 

Existing works on multilingual NMT can be categorized from different perspectives: (1) rich/low resource, whether the focused language pairs are rich/low-resource \cite{DBLP:conf/emnlp/ZophYMK16,DBLP:conf/naacl/GuHDL18,DBLP:conf/emnlp/GuWCLC18,neubig2018rapid};  (2) one-to-many/many-to-one/many-to-many translation, translation from one source language to multiple target languages \cite{DBLP:conf/acl/DongWHYW15,tan2018multilingual,tan2019multilingual}, from multiple source languages to one target language \cite{DBLP:conf/naacl/ZophK16,DBLP:journals/tacl/LeeCH17}, or from multiple source languages to multiple target languages \cite{DBLP:conf/naacl/FiratCB16,DBLP:conf/emnlp/FiratSAYC16,DBLP:journals/tacl/JohnsonSLKWCTVW17,DBLP:journals/corr/HaNW16,DBLP:conf/coling/BlackwoodBW18,he2019language}; (3) pivot translation \cite{DBLP:conf/emnlp/FiratSAYC16,DBLP:journals/tacl/JohnsonSLKWCTVW17,DBLP:journals/corr/HaNW16,cheng2017joint,chen2017teacher,leng2019unsupervised}, which bridges a source language and a target language through some pivot languages; and (4) zero-shot translation \cite{DBLP:journals/tacl/JohnsonSLKWCTVW17,DBLP:journals/corr/HaNW16}, which directly translates a source language to a target language with zero bilingual data using a multilingual model. 

Since multiple languages are involved, multilingual NMT is a very complex problem. Therefore, even if many works above have explored different aspects of multilingual NMT, there are still many open questions to investigate and answer. In this paper, we aim to achieve a deep understanding of multilingual NMT and conduct a (relatively) systematic study. In particular, we try to answer the following questions:

\textbf{Q1: Which languages should be handled together?}

Most existing works simply assume a set of languages or language pairs and use their data to train a multilingual model. There are few studies on which languages should be handled together in one multilingual model. While it is easy to get that we should handle similar languages in one model, how about dissimilar languages? Will handling dissimilar languages in one model hurt the translation accuracy of a specific language pair? Furthermore, many previous works have observed that low-resource language pairs improve in accuracy from multilingual training \cite{DBLP:journals/tacl/JohnsonSLKWCTVW17,neubig2018rapid}. How about rich-resource language pairs? 

\textbf{Q2: Does fine-tuning perform better than training a multilingual model from scratch?}

A clear advantage of multilingual NMT is that it can boost the accuracy of a low-resource language pair by leveraging bilingual training data from other (especially rich-resource) language pairs, and multiple algorithms have been proposed. \cite{DBLP:conf/emnlp/ZophYMK16,neubig2018rapid} train a model on the data of rich-resource language pairs and then fine-tune the model using the data of the target low-resource language pair. \cite{DBLP:conf/emnlp/GuWCLC18} leverage the model-agnostic meta-learning algorithm to find a better initialization for low-resource language pairs. \cite{song2019mass} leverage masked sequence to sequence pre-training for zero or low-resource languages. \cite{DBLP:conf/emnlp/PlataniosSNM18} propose a novel contextual parameter generator to generate model for low/zero-resource language pairs. Among these works, fine-tuning is the simplest one and widely used~\cite{DBLP:conf/emnlp/ZophYMK16,neubig2018rapid,DBLP:conf/emnlp/FiratSAYC16}  since it can quickly adapt a well-trained model to low-resource language pairs. Given the small amount of training data, fine-tuning a model for low-resource language pairs might be easy to overfit. An alternative solution is to train a model of low-resource language pairs together with rich-resource languages from scratch. Will training from scratch lead to better accuracy?

\textbf{Q3: Which component of a well-trained NMT model should be fine-tuned when transferring to low-resource language pairs?}

Considering limited data for low-resource language pairs, fine-tuning the whole model~\cite{DBLP:conf/emnlp/ZophYMK16,neubig2018rapid,DBLP:conf/emnlp/FiratSAYC16} may not be optimal due to overfitting. Which components of the model (i.e., encoder, decoder, attention) should be fine-tuned to ensure better accuracy. Are there any insights or guidelines to decide which component to fine-tune?

\textbf{Q4: Is pivot translation better than direct translation in low-resource settings?}

Pivot translation leverages a third language, which is usually a popular language, to bridge the translation from a source language to a target language, i.e., source$\to$pivot$\to$target. Given that previous works~\cite{cohn2007machine,DBLP:conf/acl/WuW07,utiyama2007comparison,DBLP:conf/emnlp/FiratSAYC16,DBLP:journals/tacl/JohnsonSLKWCTVW17,DBLP:journals/corr/HaNW16,cheng2017joint,chen2017teacher,leng2019unsupervised} have already shown that pivot translation is helpful when the training data for the pivot pairs are resource rich and the direct translation pair is resource poor. How does the accuracy change with respect to different amounts of direct and pivot data? Is pivot translation always better than direct translation when the direct training data is scarce? 

\textbf{Q5: What factors impact the accuracy of zero-shot translation?}

Zero-shot translation~\cite{DBLP:journals/tacl/JohnsonSLKWCTVW17,DBLP:journals/corr/HaNW16} usually refers to when there is no training data for a source$\leftrightarrow$target pair, but exists some training data for source/target$\leftrightarrow$other languages, where other languages are usually called pivot languages. Note that zero-shot translation is different from previous pivot translation in that the former one directly translates the source language to the target language in one hop using one model (i.e., a multilingual model) while the latter one translates in multiple hops (e.g., two hops, first from source to pivot and then from pivot to target) using multiple models (e.g., two models, the source-to-pivot model and the pivot-to-target model) . It is easy to understand that training a multilingual model using languages similar to the source/target languages will help zero-shot translation. Here we would like to ask one question: what other factors will impact the accuracy of zero-shot translation?

In this paper, we conduct a series of experiments on multilingual NMT, in order to analyze and answer the questions listed above. The remaining part of this paper is organized as follows. We introduce the basic experiment settings in Section~\ref{sec_exp}. We conduct detailed analyses for each individual question in Section~\ref{sec_q1}-\ref{sec_q5}. In the last section, we summarize our studies and discuss future research directions.

\section{Experimental Setup}
\label{sec_exp}
In this section, we briefly introduce the experiment settings we use to conduct the analyses and studies in this work.

\paragraph{Dataset} We use the common corpus of TED talks which contains translations between multiple languages~\cite{Ye2018WordEmbeddings}\footnote{https://github.com/neulab/word-embeddings-for-nmt}. Specifically, it consists of more than 50 languages with bilingual data between any two languages. Some language pairs have relatively large data and some languages with small data size, providing a good scenario for the multilingual and low-resource study. We also manually remove the training data from some language pairs to simulate the zero-resource setting. The detailed data statistics for the TED talks dataset are listed in the Supplementary Materials (Section 1). All the language names and its corresponding IS0-639-1 code\footnote{https://www.iso.org/iso-639-language-codes.html} can also be found in the Supplementary Materials (Section 2).

\paragraph{Model Configurations}
We use the Transformer~\cite{DBLP:conf/nips/VaswaniSPUJGKP17} as the basic NMT model structure since it achieves state-of-the-art accuracy and becomes a popular choice for recent NMT researches. We use the basic model configuration for all experiment setting unless otherwise stated. The model embedding size, hidden size and number of layers are 256, 256 and 6 respectively. For the multilingual model training, in order to give the model a sense which language it is currently processing, it is a common practice to add a special tag in the model to determine which target language to translate. Multilingual NMT usually consists of three settings: many-to-one, one-to-many and many-to-many. Here we study the many-to-one and one-to-many setting in the first three questions, since many-to-many setting can be bridged by the first two settings. We further include many-to-many setting in the last two questions about pivot and zero-shot translation. 

\paragraph{Training and Inference}
For multilingual model training, we upsample the data of each language to make all languages have the same size of data. The minibatch size is set to roughly 4096 tokens. We train the models on NVIDIA V100 GPU. We follow the default parameters of Adam optimizer~\cite{kingma2014adam} and learning rate schedule in~\cite{DBLP:conf/nips/VaswaniSPUJGKP17}. During inference, we decode with beam search and set the beam size to 6 and length penalty $\alpha=1.1$ for all the languages. We evaluate the translation quality by tokenized case sensitive BLEU~\cite{DBLP:conf/acl/PapineniRWZ02} with multi-bleu.pl\footnote{https://github.com/moses-smt/mosesdecoder/blob/ master/scripts/generic/multi-bleu.perl}. 

\section{Multilingual Model vs. Individual Model}
\label{sec_q1}

Many previous works ~\cite{DBLP:journals/tacl/JohnsonSLKWCTVW17,DBLP:journals/corr/HaNW16} simply put a set of languages together and train a single model for them. A natural question is, for a specific language pair, does a multilingual model (trained with the data from multiple language pairs) performs better than an individual model (trained with data only from this language pair)?  In this section, we study this question considering two factors: the similarity between languages and the amount of training data.


\subsection{Experiment Design}
We choose four language pairs: En$\leftrightarrow$De and En$\leftrightarrow$Fr, two rich-resource language pairs with 168K and 192K bilingual sentence pairs respectively, and En$\leftrightarrow$Nb and En$\leftrightarrow$Da, two  low-resource language pairs with 16K and 45K bilingual sentences pairs respectively. The four languages come from the Germanic (De, Nb, Da) and Romance (Fr) branch of the Indo-European language family. We first train an individual model (\textit{Individual}) for each pair using only the bilingual data from that language pair. Then for each language pair, we train two multilingual models: (1) the \textit{Multi-sim} model, trained with the bilingual data from both the original language pair and 5 other similar language pairs in the similar (Germanic/Romance) language branches from the same language family (Indo-European), i.e., En$\leftrightarrow$Es/Nl/It/Pt/Ro with 196K/184K/205K/185K/180K bilingual sentence pairs, and (2) the \textit{Multi-dissim} model, trained with the bilingual data from both the original language pair and 5 other dissimilar language pairs from different language branches or language families, i.e., En$\leftrightarrow$Ar/Bg/Ja/Ru/Zh with 214K/174K/204K/208K/200K bilingual sentence pairs.

\begin{table}[h]
\centering
\begin{tabular}{lccc}
\toprule
 & \textit{Individual}  &  \textit{Multi-sim}  & \textit{Multi-dissim} \\
\midrule
En-De          & 28.82  & 27.54 & 26.56 \\
De-En          & 34.61  & 34.15 & 33.40 \\
\midrule
En-Fr & 41.94 & 41.03 & 36.64  \\
Fr-En & 40.10 & 39.65 & 37.39 \\
\toprule
En-Nb & 25.34  & 36.35 & 34.52 \\
Nb-En & 26.84  & 43.52 & 41.40 \\
\midrule
En-Da & 22.68 & 33.42 & 31.15 \\
Da-En & 26.09 & 42.28 & 41.16 \\
\bottomrule
\end{tabular}
\vspace{10pt}
\caption{The BLEU scores on rich-resource language pairs (En$\leftrightarrow$De and En$\leftrightarrow$Fr) and low-resource language pairs (En$\leftrightarrow$Nb and En$\leftrightarrow$Da) when training alone (\textit{Individual}) and  training with other similar/dissimilar languages (\textit{Multi-sim}/\textit{Multi-dissim}).}
\label{table_different_lan}
\end{table}

\begin{table}[h]
\centering
\small
\begin{tabular}{lccc}
\toprule
 & \textit{Individual}  &  \textit{Multi-sim}  & \textit{Multi-dissim} \\
\midrule
En-De (\textit{larger}) & 29.32  & 29.75 & 28.79 \\
De-En (\textit{larger}) & 34.98  & 35.17 & 34.73 \\
\midrule
En-Nb (\textit{smaller}) & 24.32 & 30.25 & 28.91 \\
Nb-En (\textit{smaller}) & 25.49 & 40.20 & 38.07 \\
\bottomrule
\end{tabular}
\vspace{10pt}
\caption{The BLEU scores on rich-resource language (En$\leftrightarrow$De) and low-resource language (En$\leftrightarrow$Nb) when training alone (\textit{Individual}) and  training with other similar/dissimilar languages (\textit{Multi-sim}/\textit{Multi-dissim}). \textit{larger}/\textit{smaller} mean we increase the mode size for the rich-resource language and decrease the model size for the low-resource language compared with that in Table~\ref{table_different_lan}.}
\label{table_different_lan_larger_smaller}
\end{table}

\subsection{Results and Analysis} 
The results are shown in Table~\ref{table_different_lan}. It can be seen that the accuracy of multilingual models for both En$\to$De and De$\to$En is worse than the individual models, no matter training with similar languages or dissimilar languages. Furthermore, the BLEU score of the \textit{Multi-dissim} model drops more than that of the \textit{Multi-sim} model. We have similar observations on En$\to$Fr and Fr$\to$En, i.e.,  dissimilar languages have larger negative impacts than similar languages. 

For the low-resource language En$\to$Nb and Nb$\to$En, no matter training with similar or dissimilar languages, the accuracy can always be improved. This shows that low-resource language pair benefits from multilingual training. Furthermore, training with similar languages brings more accuracy gain than training with dissimilar languages, which shows that similar languages help more with multilingual training. Similar patterns can be observed from En$\to$Da and Da$\to$En. 

So far, we observe that low-resource language pairs benefit from multilingual training, while rich-resource language pairs are hurt from multilingual training, and more hurts when trained with dissimilar language pairs. Our hypothesis for this phenomenon is that for a rich-resource language pair, its training data well matches its model capacity, and further adding more data from other (even similar) language pairs goes beyond its model capacity therefore leading to accuracy drop; in contrast, for a low-resource language pair, its limited training data under match its model capacity (with same structure and number of parameters as the rich-resource language pair), thus introducing more training data from other language pairs improves translation accuracy~\cite{DBLP:conf/emnlp/ZophYMK16,dabre2017empirical}. To verify our hypothesis, we conduct two additional experiments, one to enlarge the model size for the rich-resource pair (i.e., from 6 layers and 256 hidden dimensions to 6 layers and 512 hidden dimensions) and the other one is to reduce the model size for the low-resource language pair (i.e., from 6 layers and 256 hidden dimensions to 6 layers and 128 hidden dimensions). As shown in Table~\ref{table_different_lan_larger_smaller}, with a larger model compared with that in Table~\ref{table_different_lan}, the rich-resource language pair also benefits more from multilingual training, and with a smaller model compared with that in Table~\ref{table_different_lan}, the low-resource language pair benefits less from multilingual training. 


As a summary, our studies show that (1) the accuracy of the rich-resource language becomes worse when training with multiple languages while slightly improving with bigger model capacity, (2) low-resource language pair benefits more from multilingual training, and (3) multilingual training with similar languages brings more improvement than dissimilar languages.


\section{Fine-tuning vs. From Scratch}
\label{sec_q2}

In the previous section, we show that low-resource language pairs benefit from multilingual training. Different strategies have been adopted to boost low-resource pairs by multilingual training, including fine-tuning, which first trains a multilingual model using other (rich-resource) language pairs and then fine-tunes the model using the data from the target language pair (\textit{Multi-fine-tune}), and multilingual training from scratch, which directly trains a multilingual model from scratch by mixing data from the target low-resource language pair and other language pairs (\textit{Multi-scratch}). \cite{neubig2018rapid} have also found that both \textit{Multi-fine-tune} and \textit{Multi-scratch} (corresponding to the cold-start and warm-start setting in their paper) could be beneficial. In this section, we study which strategy is more effective in one-to-many and many-to-one setting.

\subsection{Experiment Design} 
We consider three low-resource language pairs as our targets: En$\leftrightarrow$Nb, En$\leftrightarrow$Da and En$\leftrightarrow$Ka, each with 16K/45K/13K bilingual sentence pairs. For the three pairs, we leverage En$\leftrightarrow$Es/Nl/It/Pt/Ro for multilingual training, and the number of bilingual sentence pairs are 196K/184K/205K/185K/180K for these pairs. We use the transformer model with the default setting in our paper: 6-layer encoder and decoder, 256 embedding dimension and 256 hidden dimension. We also train a model using the data only from the target language pair (\textit{Individual}) as a reference. We study both the one-to-many and many-to-one settings.

\begin{table}[h]
\centering
\begin{tabular}{cccc}
\toprule
&  \textit{Individual} & \textit{Multi-scratch} & \textit{Multi-fine-tune} \\
\midrule
En$\to$Nb  & 25.34 & 36.95 & \textbf{38.36} \\
En$\to$Da & 22.68 & 33.42 & \textbf{35.38} \\
En$\to$Ka & 7.23  & 11.80 & \textbf{13.08}  \\
\midrule
Nb$\to$En & 26.84  & \textbf{44.41} & 43.51\\
Da$\to$En & 26.09 & \textbf{42.28} & 40.59 \\
Ka$\to$En & 11.01 & \textbf{20.75} & 15.24  \\
\bottomrule
\end{tabular}
\vspace{10pt}
\caption{ BLEU scores comparisons between multilingual training from scratch and multilingual training with fine-tuning for one-to-many setting (En$\to$Nb/Da/Ka) and many-to-one setting (Nb/Da/Ka$\to$En).}
\label{table_fine-tune}
\end{table}

\subsection{Results and Analysis} 
The results are shown in Table~\ref{table_fine-tune}. It is clear that both \textit{Multi-scratch} and \textit{Multi-fine-tune} outperform the individual model for both language pairs and in both settings. \textit{Multi-fine-tune} outperforms \textit{Multi-scratch} on En$\to$Nb/Da/Ka (one-to-many setting), while \textit{Multi-scratch} is better on Nb/Da/Ka$\to$En (many-to-one setting). This is consistent with our intuition that the encoder in NMT model is usually in charge of extracting the source sentence representation for the decoder, and the encoder in one-to-many setting handles the common source language (English), which is well trained and ready to be tuned for new target language. However, the encoder in the many-to-one setting should be adapted, and only fine-tuning results in worse performance than training from scratch.  

As a summary, our studies suggest that for multilingual translation, fine-tuning is better for the one-to-many setting while training from scratch performs better in the many-to-one setting.

\section{Which Components to Fine-tune?}
\label{sec_q3}
In the previous section, we show that multilingual training with fine-tuning can better boost the accuracy of low-resource language pairs for the one-to-many setting. Considering the limited data for low-resource language pairs, fine-tuning the whole model~\cite{DBLP:conf/emnlp/ZophYMK16,DBLP:conf/emnlp/FiratSAYC16,neubig2018rapid} may not be a good choice. Therefore, a follow-up question is which component of the model we should fine-tune. ~\cite{DBLP:conf/emnlp/ZophYMK16,sachan2018parameter} have also studied the sharing of different components of a multilingual model. However, they either focus on high-level granularity of the whole encoder and decoder, or the lower-level one of the self-attention and feed-forward structures in Transformer model~\cite{DBLP:conf/nips/VaswaniSPUJGKP17}. In this section, we try to analyze with a suitable granularity for the comparison between one-to-many and many-to-one setting.

\subsection{Experiment Design} 
We split the transformer model (6-layer encoder/decoder with attention) into 6 parts: (1) word embedding, which includes the source/target embedding as well as the output softmax matrix; (2) the bottom 3 layers of the encoder (denoted as \textit{encoder\_bottom}); (3) the top 3 layers of the encoder (\textit{encoder\_top}); (4) the attention between encoder and decoder (\textit{attention}); (5) the bottom 3 layers of the decoder (\textit{decoder\_bottom}); (6) the top 3 layers of the decoder (\textit{decoder\_top}). 

Same as the previous section, we choose En$\to$Nb and En$\to$Ka as our target language pairs, which are of low resource. We first train the model on En$\to$Es/Fr/It/Pt/Ro (one-to-many setting) and then fine-tune each component of the model on En$\to$Nb and En$\to$Ka respectively. The word embedding part is fine-tuned in all the experiments in this section so as to adapt the translation model to the target languages (i.e., Nb and Ka). Therefore, we get 5 component-wise fine-tuned models. The scale of the training data for each language pair and the model configuration are the same as the previous section.

\begin{table}[h]
\centering
\begin{tabular}{lll}
\toprule
Settings           & En$\to$Nb    & En$\to$Ka  \\ 
\midrule
\textit{encoder\_bottom}    & 35.06        & 10.38           \\
\textit{encoder\_top}       & 37.53        & 11.34           \\
\textit{attention}          & 37.62        & 12.11           \\
\textit{decoder\_bottom}    & 37.60        & 12.46           \\
\textit{decoder\_top}       & \textbf{38.20}        & \textbf{13.06}           \\
\bottomrule
\end{tabular}
\caption{Results of fine-tuning different components in one-to-many setting.}
\label{component-exp-table-enxx}
\end{table}

\begin{table}[h]
\centering
\begin{tabular}{lll}
\toprule
Settings           & Nb$\to$En    & Ka$\to$En  \\ 
\midrule
\textit{encoder\_bottom}    & \textbf{43.29}        & \textbf{15.15}      \\
\textit{encoder\_top}       & 42.87        & 14.77      \\
\textit{attention}          & 41.50        & 12.70      \\
\textit{decoder\_bottom}    & 40.42        & 11.81      \\
\textit{decoder\_top}       & 39.48        & 11.65      \\
\bottomrule
\end{tabular}
\caption{Results of fine-tuning different components in many-to-one setting.}
\label{component-exp-table-xxen}
\end{table}

\subsection{Results and Analysis} 
The results of fine-tuning 5 different model components are shown in Table~\ref{component-exp-table-enxx}. It can be seen that fine-tuning \textit{decoder\_top} in the one-to-many setting results in higher accuracy than other components. Fine-tuning \textit{decoder\_top} can nearly match the accuracy of fine-tuning all components (38.36 and 13.08 for En$\to$Nb and En$\to$Ka as shown in Table~\ref{table_fine-tune}), which demonstrates that \textit{decoder\_top} is more important to characterize the language specific information in the one-to-many setting. This is consistent with our intuition: in the one-to-many setting, the encoder in the pre-training stage and the fine-tuning stage handles the same input language (here is English), while the decoder handles different languages in pre-training and fine-tuning, and thus we need to fine-tune the decoder, especially the top layers of the decoder which are close to the final output, to better adapt to the output language of the target pair. To further verify our intuition, we conduct the same experiments for the many-to-one setting. As shown in Table~\ref{component-exp-table-xxen}, fine-tuning the encoder, especially the bottom layers of the encoder, leads to better accuracy, since in this many-to-one setting, the encoder needs to handle diverse input languages.

\section{Pivot Translation vs. Direct Translation}
\label{sec_q4}

For a low-resource language pair, pivot translation (e.g., translation from the source language to a pivot language like English and then from the pivot language to the target language) is widely adopted in real-world systems ~\cite{DBLP:journals/tacl/JohnsonSLKWCTVW17,DBLP:journals/corr/HaNW16,cheng2017joint,chen2017teacher}. Here we investigate whether and when pivot translation is better than direct translation.

\subsection{Experiment Design} 
We take English as the pivot language considering its global popularity. We choose 17 other languages from Indio-European family, including three language branches: Germanic (4 languages), Romance (5 languages), and Slavic (8 languages). We train $17\times16=272$ individual models for direct translation, each model using only the bilingual data for one language pair. We then train extra $17\times2=34$ individual models on the 17 languages$\leftrightarrow$English, for the pivot translation between any two of the 17 languages. 

\begin{table*}[h]
\centering
\begin{tabular}{cccccccc}
\toprule
Source & Target & Pivot BLEU & Direct BLEU &  $\Delta$  & Source$\to$Pivot Data & Pivot$\to$Target Data & Direct Data \\
\midrule
Bs & Mk & 19.43 & 21.51 &-2.08  & 5661  & 25335 & 1692   \\
Da & Sv & 26.79 & 28.85 &-2.06  & 44925 & 56646 & 17098 \\
Sk & Mk & 16.71 & 18.40 & -1.69 & 61454 & 25335 & 11096 \\
Nb & Sv & 23.68 & 25.11 & -1.43 & 15819 & 56646 & 7109 \\
\bottomrule
\end{tabular}
\caption{Some examples that direct translation is better than pivot translation through English while the direct training data is less than the pivot training data. $\Delta$ means the BLEU gap (the BLUE score of pivot translation minus that of direct translation). The last three columns list the number of training sentence pairs for source$\to$pivot translation, pivot$\to$target translation, and direct translation.} 
\label{data_pivot_direct_compare}
\end{table*}

\subsection{Results and Analysis}
Intuitively, there are two major factors impacting the accuracy of direction translation and pivot translation: (1) the similarity between the source language and the target language, and (2) the size of the training data of the source-target language pair and source/target-pivot language pair. Therefore, we compare the results of direct translation and pivot translation from two aspects.

\begin{figure}[h]
    \centering
        \includegraphics[width=0.45\textwidth]{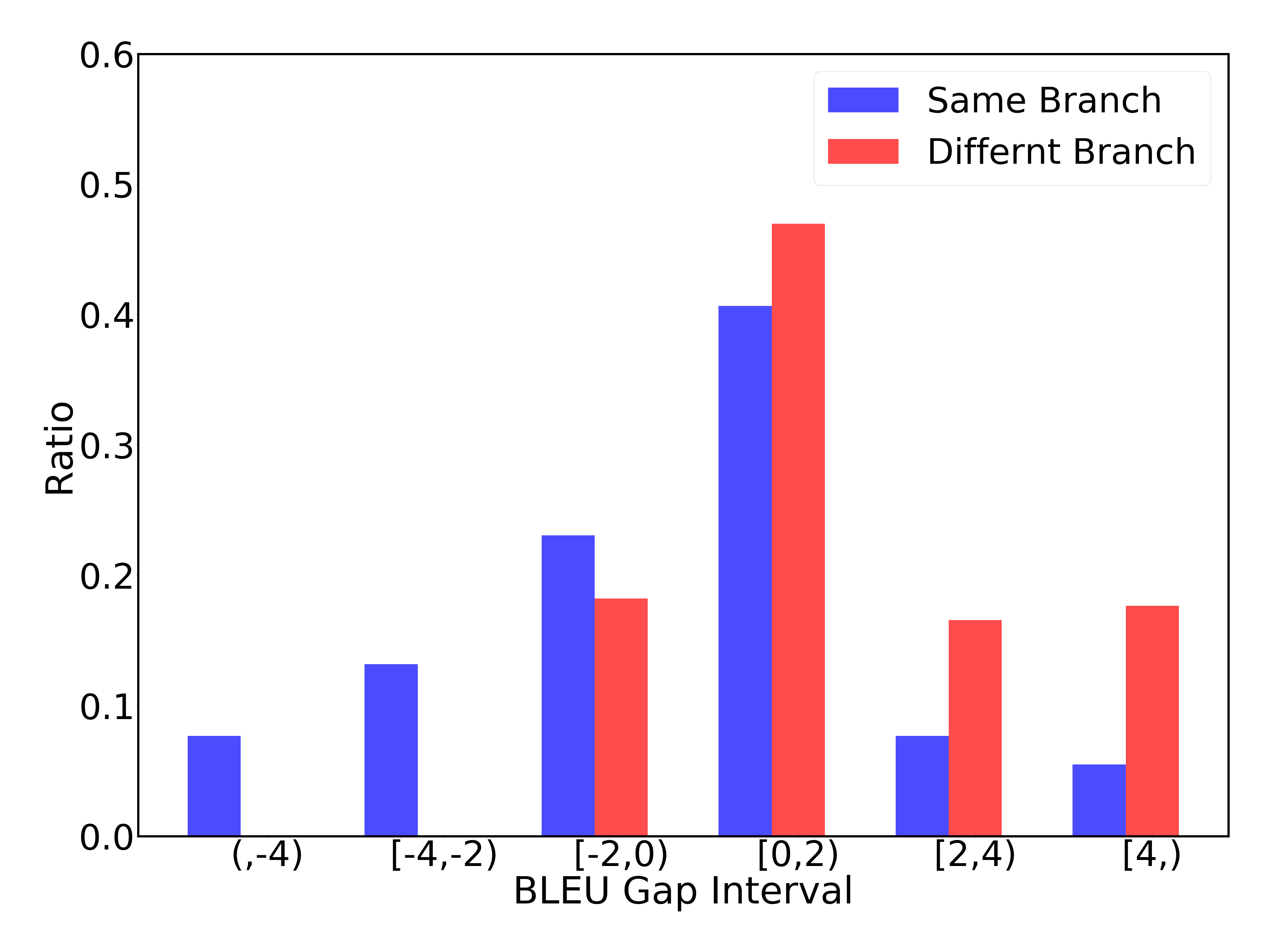}
        \vspace{10pt}
     \caption{The ratio of language pairs in different BLEU gap intervals (the BLEU gap represents the BLUE score of pivot translation minus that of direct translation) for the source and target language in the same branch and in different branch.} 
   \label{bleu_gap_same_diff_branch}
\end{figure}

We first check how the similarity between source and target language (within/across language branches) impacts the performance of direct translation and pivot translation. As shown in Figure~\ref{bleu_gap_same_diff_branch}, for the source and target language in different branches, pivot translation outperforms direct translation on 81.7\% of language pairs, while for the source and target language in the same branch, the distribution shifts left, indicating the number of language pairs on which direct translation outperforms pivot translation increases\footnote{There are nearly 44.9\% direct translations work better for the language in the same branch. Note that 44.9\% is already a high ratio considering that the training data for the direct translation is smaller than the pivot translation for almost all the language pairs in our dataset.}.  

\begin{figure}[!h]
    \begin{subfigure}{0.23\textwidth} 
        \includegraphics[width=\textwidth]{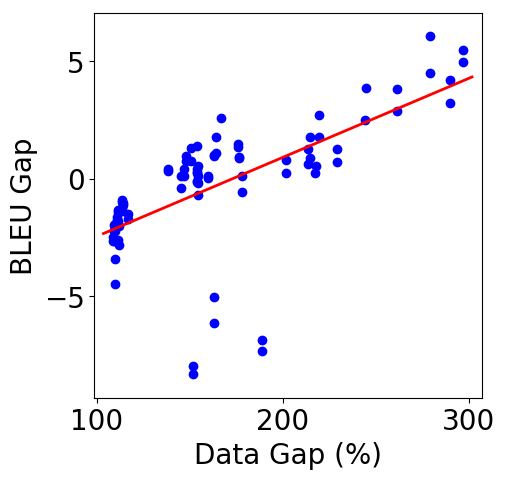}
        \vspace{4pt}
        \caption{Source and target language in the same branch} 
        \label{bleu_gap_data_size_same}
    \end{subfigure}
      \begin{subfigure}{0.235\textwidth} 
        \includegraphics[width=\textwidth]{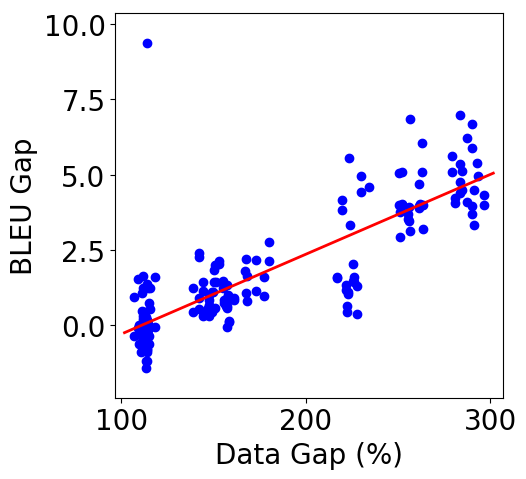}
        \vspace{1pt}
        \caption{Source and target language in different branches} 
        \label{bleu_gap_data_size_diff}
    \end{subfigure}  
    \vspace{1pt}
     \caption{ The scatter plot of the BLEU gap and the data ratio (the data size of the pivot translation over that of the direct translation) on the language pairs in the same branch (Figure a) and different branches (Figure b). The red line represents the first order curve fitting.} 
   \label{bleu_gap_data_size}
\end{figure}

We then check how the size of training data impacts the performance of direct translation and pivot translation. The X axis of Figure~\ref{bleu_gap_data_size} shows the ratio of the training data size of pivot translation over that of direct translation. Here we simply count the average number of bilingual sentence pairs of source-to-pivot language pair and pivot-to-target language pair and treat it as the size of pivot translation. The training data size of direct translation is the number of bilingual sentence pairs of the source-to-target language pair. The Y axis shows the accuracy gap, i.e., BLUE of pivot translation minus that of direct translation. From the figures, we can see that the more data the pivot translation has, the larger accuracy improvement it will achieve, no matter whether the source and target languages are in the same language branch (Figure~\ref{bleu_gap_data_size_same}) or not (Figure~\ref{bleu_gap_data_size_diff}). 

It can be seen from the left part of Figure~\ref{bleu_gap_data_size_same} that the BLEU score of the pivot translation is lower than that of the direct translation (BLEU gap is lower than zero), even if the pivot data is more than the direct data (data gap is bigger than 100\%). We show some examples from this part in Table~\ref{data_pivot_direct_compare}. For these language pairs, the source and target languages are very similar, such as Bs (Bosnian) and Mk (Macedonian), which are both in the South Slavic branch and both adopt the Serbian Cyrillic alphabet. Another case is that Da (Danish), Nb (Norwegian Bokmal) and Sv (Swedish) are very close as they are all Scandinavian in North Germanic branch.

As a summary, the observations in this section is quite consistent with our intuition: (1) When the source and target languages are similar (e.g., in the same language branch), direct translation performs better on a large part of language pairs, while pivot translation performs better for the majority of dissimilar languages; (2) the improvement of pivot translation over direct translation positively correlates with the ratio of the training data size. An interesting point is that direct translation of very similar source/target languages could perform better than pivot translation, even if pivot translation may have a larger scale of training data. This motivates us to rethink about the widely used pivot translation. In some situations (see Table \ref{data_pivot_direct_compare}), direct translation with less training data is already better than resource-consuming pivot translation with more training data.

\section{Zero-Shot Translation}
\label{sec_q5}

To translate a language pair with zero bilingual data, one can leverage other language pairs. A simple approach is to choose a pivot language, train a model for source-to-pivot translation and a model for pivot-to-target translation, and then translate a source sentence to the target language through the pivot language using the two models. One issue with pivot translation is that it takes twice the translation time of direct translation, since it conducts two steps of translation. Zero-shot translation, which trains a model using multiple language pairs and then directly translates a source sentence to a target sentence, can avoid the latency issue of pivot translation and thus has been studied in multilingual NMT~\cite{DBLP:journals/tacl/JohnsonSLKWCTVW17}. It is well known that adding similar languages would benefit for the zero-shot translation. Here we study an even more interesting phenomenon that the diversity of language itself would help on the zero-shot translation.


\subsection{Experiment Design} 
We choose two languages Pl and Pt in this section, assuming that there are no bilingual training data between them. First, we train a multilingual model using the bilingual data for the 2 language pairs (i.e., Pl/Pt$\leftrightarrow$English), then use this model to directly translate between Pl and Pt (zero-shot). We denote this model as \emph{Zero-2}. Second, we randomly add 10 more languages (Be/Bn/Hi/Ja/Kk/Ku/Ro/Ta/Tr/Zh) to train a second multiple model, using the bilingual data for 12 languages$\leftrightarrow$English. We denote this model as \emph{Zero-12}. Third, we randomly add 20 more languages (the original 10 languages: Be/Bn/Hi/Ja/Kk/Ku/Ro/Ta/Tr/Zh, and 10 other languages: De/Eo/Et/Eu/Fa/Gl/He/Hr/Sv/Uk) over the model \emph{Zero-2} to train a second multilingual model, using the bilingual data for 22 languages$\leftrightarrow$English. We denote this model as \emph{Zero-22}. To ensure the size of bilingual training sentence pairs comparable between the second and third models, we only use half of the bilingual data of the 20 languages for \emph{Zero-22}. As we choose the languages at random, the total training data size of \emph{Zero-22} is roughly the same as that of \emph{Zero-12}.

\subsection{Results and Analysis} 

\begin{figure}[h]
    \centering
        \includegraphics[width=0.45\textwidth]{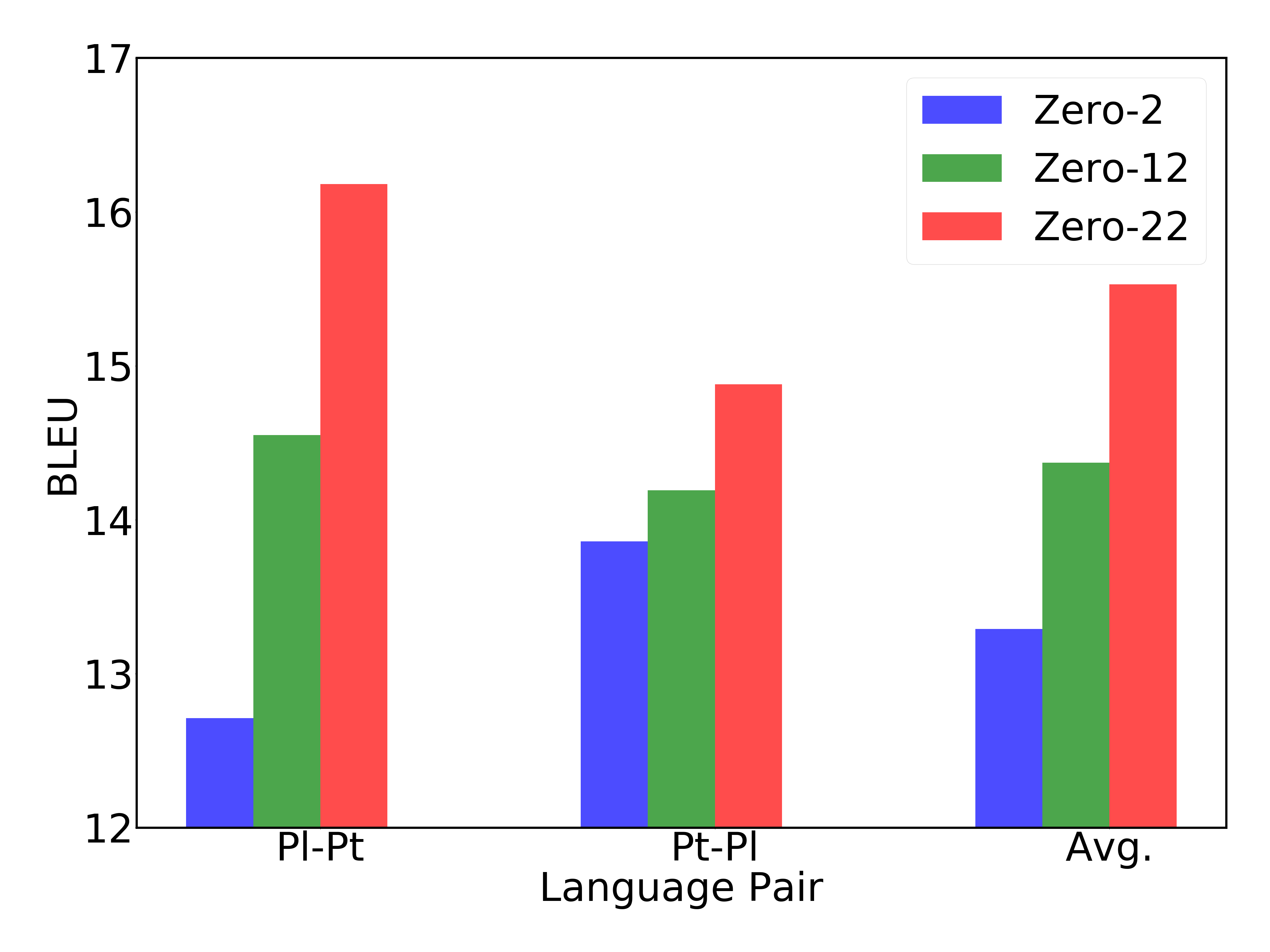}
     \caption{The BLEU scores comparison between different zero-shot translation settings: \textit{Zero-2}, \textit{Zero-12} and \textit{Zero-22}. ``Avg." means the average of the BLEU scores for Pl$\to$Pt and Pt$\to$Pl.} 
   \label{zero_shot_compare}
\end{figure}

The results of the three multilingual models for zero-shot translation are shown in Figure~\ref{zero_shot_compare}. It can be seen that \emph{Zero-12} is much better than \emph{Zero-2}, because \emph{Zero-12} uses more data from other languages for training. What is interesting is that \emph{Zero-22} is much better than \emph{Zero-12}. Note that the size of the training data of the two models are roughly the same, and thus the improvement of \emph{Zero-22} over \emph{Zero-12} mainly comes from the introduction of more diverse languages for model training. An intuitive explanation is that by introducing more diverse languages into training, a multilingual model can learn the implicit universal language representations and consequently improve zero-shot translation. 

\section{Summary and Future Work}
\label{sec_discuss}
In this paper, we have conducted a comprehensive study on multilingual neural machine translation to gain a deep understanding of this direction. Note that we conduct a variety of experiments on dozens of languages to study multilingual NMT. We carefully formulate this paper in order to demonstrate the results in a clear logic, without showing the results of all languages that are in the same phenomenon. We list several key takeaways from our studies. 

\begin{itemize}
\item Low-resource language pairs benefit more from multilingual training, while rich-resource pairs may be hurt by limited model capacity. They get more benefits from training with similar languages than dissimilar languages.

\item Multilingual training with fine-tuning  performs  better  for  the one-to-many  setting,  while  training  from  scratch performs better in the many-to-one setting.

\item Fine-tuning top layers of the decoder of the model works better for the one-to-many setting, while fine-tuning bottom layers of the encoder works better for the many-to-one setting.

\item Direct translation often performs better for a pair with similar source and target languages, even if the direct translation has fewer data than the pivot translation, while pivot translation works better especially for a pair with dissimilar source and target languages. 

\item Zero-shot translation benefits from multilingual training with more and diverse languages.
\end{itemize}

Based on the studies in this work, we point out several future research topics that can further improve multilingual NMT:
\begin{itemize}
\item How can we better cluster languages according to their similarities to decide which languages should be trained together in one multilingual model? 
\item How to maintain the accuracy of a multilingual model when handling diverse languages with limited model capacity? 
\item Can we select a better pivot language or even multiple pivot languages for long-distance language pairs? 
\item Can we accurately model/evaluate the similarity between languages so as to choose better languages to boost the accuracy of zero-shot translation?
\end{itemize}

\bibliography{aaai}
\bibliographystyle{aaai}

\end{document}